\documentclass{article}

    \usepackage[preprint]{neurips_2024}

\usepackage[utf8]{inputenc} 
\usepackage[T1]{fontenc}    
\usepackage{hyperref}       
\usepackage{url}            
\usepackage{booktabs}       
\usepackage{amsfonts}       
\usepackage{nicefrac}       
\usepackage{microtype}      
\usepackage{xcolor}         
\usepackage{multirow}
\usepackage{paralist}
\usepackage{graphicx}
\usepackage{subcaption}
\usepackage{amsmath}
\usepackage{makecell}
\usepackage{algorithm}
\usepackage{algorithmic}
\usepackage{kotex}

\usepackage{array}
\newcolumntype{L}[1]{>{\raggedright\let\newline\\\arraybackslash\hspace{0pt}}m{#1}}
\newcolumntype{C}[1]{>{\centering\let\newline\\\arraybackslash\hspace{0pt}}m{#1}}
\newcolumntype{R}[1]{>{\raggedleft\let\newline\\\arraybackslash\hspace{0pt}}m{#1}}
\newcolumntype{Z}[1]{>{\let\newline\\\arraybackslash\hspace{0pt}}m{#1}}
\newcolumntype{P}[1]{>{\centering\arraybackslash}p{#1}}
\newcolumntype{X}[1]{>{\raggedright\arraybackslash}p{#1}}

\def\x{\mathbf{x}}
\def\y{\mathbf{y}}
\def\w{\mathbf{w}}
\def\e{\mathbf{e}}
\def\LR{\mathrm{LR}}
\def\CLIP{\mathrm{CLIP}}
\def\img{\mathrm{img}}
\def\tok{\mathrm{tok}}
\def\res{\mathrm{resid}}

\title{Improving Interpretability and Robustness for the Detection of AI-Generated Images}

\author{
  Tatiana Gaintseva$^{1}$, Laida Kushnareva$^{2}$, \\ German Magai$^{3,4}$, Irina Piontkovskaya$^{2}$, \\ Sergey Nikolenko$^{7}$, Marting Benning$^{8}$ \\
 Serguei Barannikov$^{5,6}$, Gregory Slabaugh$^{1}$,
 ,  
 \\
  \\ \textsuperscript{1} Digital Environment Research Institute, Queen Mary University of London, UK;   
  \\ \textsuperscript{2} AI Foundation and Algorithm Lab, Russia;
  \\ \textsuperscript{3} HSE University, Russia;  \textsuperscript{4} Noeon Research, Japan; \\ \textsuperscript{5} Skolkovo Institute of Science and Technology, Russia;
  \\ \textsuperscript{6} CNRS, Université Paris Cité, France;
\\\textsuperscript{7} St. Petersburg Department of the Steklov Institute of Mathematics, Russia \\
\textsuperscript{8} Department of Computer Science, 
University College London, UK\\
\\
\small{
    \textbf{Correspondence:} \href{mailto:t.gaintseva@qmul.ac.uk}{t.gaintseva@qmul.ac.uk}}
}

\begin{document}

\maketitle

\begin{abstract}
  With growing abilities of generative models, artificial content detection becomes an increasingly important and difficult task. However, all popular approaches to this prсoblem suffer from poor generalization across domains and generative models. In this work, we focus on the robustness of AI-generated image (AIGI) detectors. We analyze existing state-of-the-art AIGI detection methods based on frozen CLIP embeddings and show how to interpret them, shedding light on how images produced by various AI generators differ from real ones. Next we propose two ways to improve robustness: based on removing harmful components of the embedding vector and based on selecting the best performing attention heads in the image encoder model. Our methods increase the mean out-of-distribution (OOD) classification score by up to 6$\%$ for cross-model transfer. We also propose a new dataset for AIGI detection and use it in our evaluation; we believe this dataset will help boost further research. The dataset and code are provided as a supplement.
\end{abstract}

\section{Introduction}

The proliferation of generative AI has led to an explosion in AI-generated content. Large language models (LLMs) closely mimic human writing, while image generation models create increasingly realistic images with detailed control over their features. This surge in AI-generated content becomes a major challenge for AI safety and raises concerns about potential misuse, leading to the \emph{artificial content detection} problem: has a given image or other content been generated by an AI model or a human?
Detection methods for artificial content can be divided into \emph{score-based} and \emph{classifier-based}. The former identify and measure specific distinguishing features, e.g., semantic inconsistencies in images \citep{farid2022lighting} or unique ``fingerprints'' invisible to the human eye \citep{yu2019attributing}. In the text domain, \cite{gehrmann2019gltr} found statistical artifacts in LLM-generated text,
while other works measured the perplexity from another LM \citep{solaiman2019release}, curvature of the probability function \citep{mitchell2023detectgpt}, and the intrinsic dimensionality of contextualized representations \citep{tulchinskii2023intrinsic}. However, score-based methods often depend on specific generators or semantic domains, and 
known traces may be easy to remove through common transformations such as resizing or compression for images or paraphrasing for text \citep{gragnaniello2021gan,krishna2023paraphrasing}. A notable exception is the intrinsic dimension of text, which \cite{tulchinskii2023intrinsic} showed to be robust to domain transfer and paraphrasing, but its in-domain detection quality is modest.

Supervised classification methods do not require prior knowledge of the generator and can support multiple data sources. However, they typically perform poorly when transferring to unseen domains and generators. For example, \cite{corvi2023detection} propose an approach that performs nearly perfectly on GAN-generated images but drops to random guessing on diffusion-based generation. The choice of training data, including both artificial and real content, as well as appropriate data augmentations, is crucial for successful out-of-domain transfer. Unfortunately, it is often impossible to predict whether a classifier trained on a specific dataset will generalize well to new, unseen generators and data sources.
%
While features may exist that distinguish between natural and artificial subsets in the training set, classifiers often extract dataset-specific spurious differences, leading to poor generalization.
In this work, we focus on the detection of AI-generated images (AIGI), aiming to improve the robustness of supervised classification approaches. We propose several methods to remove unnecessary information from the classifier, thereby reducing overfitting and improving generalization capabilities.
Below, Section~\ref{sec:related} surveys related work, Section~\ref{sec:data} introduces the data, Section~\ref{sec:methods} shows proposed methods, Section~\ref{sec:eval} presents our experimental evaluation, and Section~\ref{sec:concl} concludes the paper.

\section{Related Work}\label{sec:related}

\begin{figure}[!t]\centering\setlength{\tabcolsep}{2pt}
\begin{tabular}{cc}
\includegraphics[width=.46\linewidth]{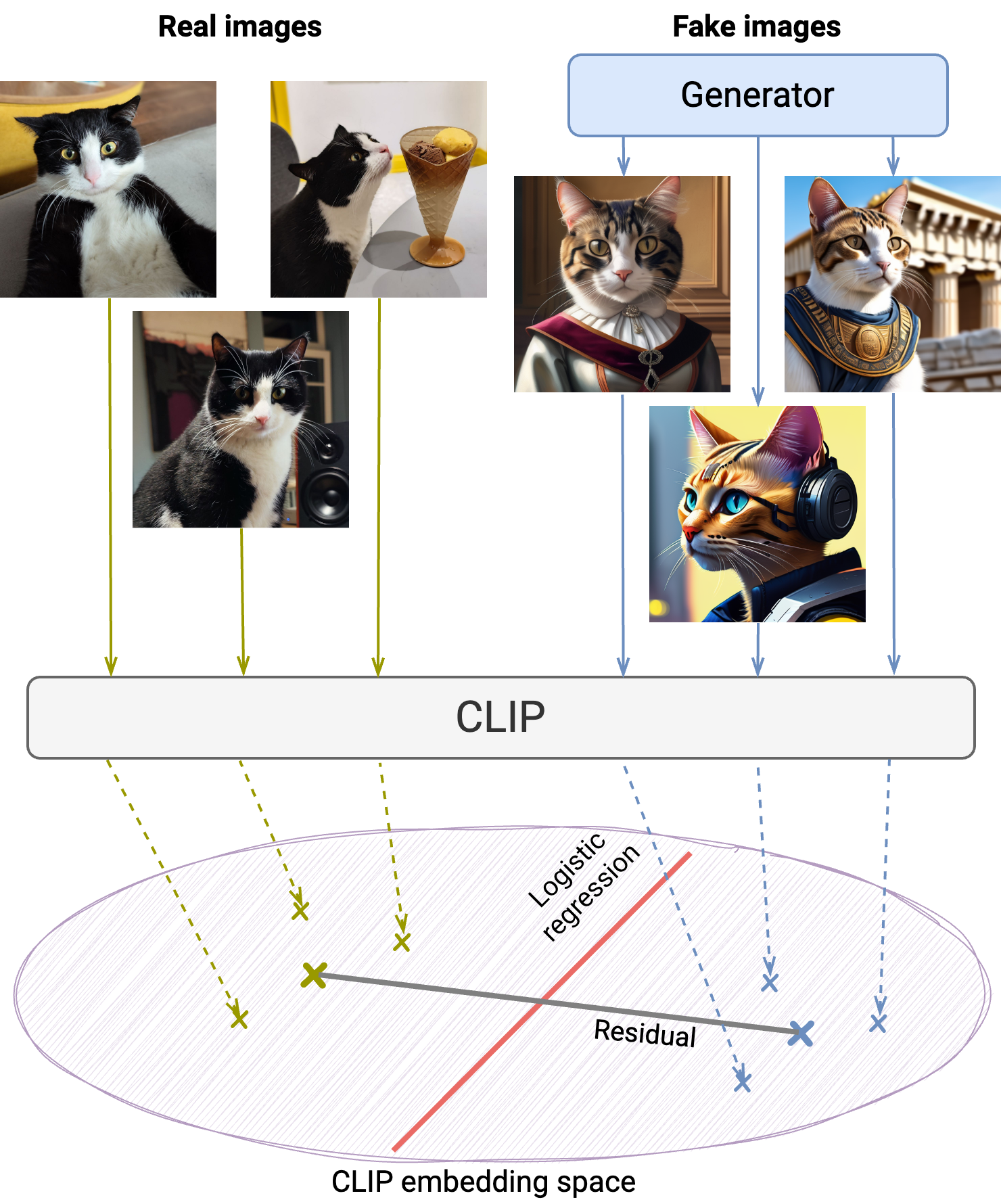} &
\includegraphics[width=.5\linewidth]{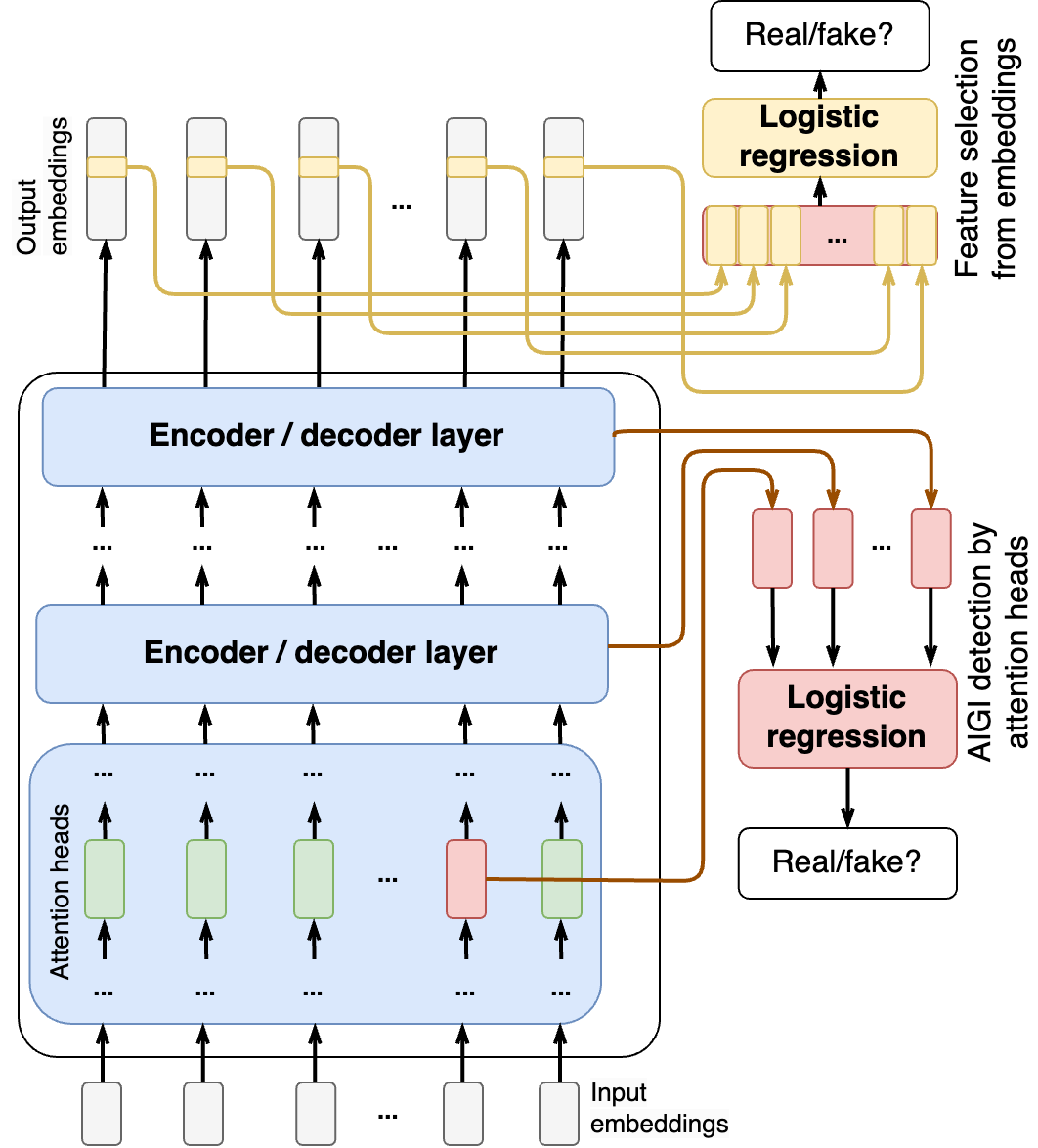}
\\ (a) & (b) 
\end{tabular}

\caption{AIGI detection: (a) CLIP embedding space; (b) attention heads and feature selection.}\label{fig:method}
\end{figure}


\textbf{Components of Transformer-based embeddings}.
Several recent works extract useful features from the geometry of inner representations or parameter spaces of large-scale neural models. One line of work considers \emph{outlier dimensions} in embedding spaces
that have an unusually high variance and/or mean value: \cite{kovaleva-etal-2021-bert} studied how outlier dimensions emerge and the effects of disabling them after training, \cite{luo-etal-2021-positional} related them to positional embeddings and influence on word-in-context tasks, \cite{timkey-van-schijndel-2021-bark} studied their influence on the quality of representations, and \cite{puccetti-etal-2022-outlier} related them to the shapes of attention maps and token frequencies.
Other works focus on probing the embeddings of a Transformer-based model: \cite{jawahar-etal-2019-bert} study LLM embeddings for language structure information, \cite{conneau-etal-2018-cram}, for semantic and syntactic features, also introducing a comprehensive selection of probing tasks. \cite{lewis2022does} probed ViT as well, although this direction is not as developed as for LLMs.
However, there is little research on how embedding dimensions of Transformer-based models influence the quality of these models; in this work we start filling this gap.

\textbf{Interpreting Vision Transformers}.
Interpretation of individual attention heads or layers is a trending research topic for LLMs. \cite{Kovaleva2019RevealingTD} proved overparametrization for BERT by pruning some heads, and \cite{Michel2019AreSH} showed that most heads can be removed at test time. Other works focused on attention head specializations \citep{clark-etal-2019-bert} and functional roles \citep{pande2021heads}. \cite{jawahar-etal-2019-bert} showed that in BERT-like models important information can be distributed across different layers.
For Vision Transformers (ViT) \citep{dosovitskiyB0WZ21}, 
\cite{chefer2021transformer} introduced a new interpretability method that computes relevance scores via the deep Taylor decomposition, providing more accurate and class-specific visual explanations.
\cite{gandelsman2024interpreting} turn to multimodal foundational models such as CLIP \citep{RadfordKHRGASAM21} and present a method for interpreting CLIP image representations by decomposing them into text-explainable components associated with specific heads attention and image locations, improving the interpretability and performance of the model.
%
%
A separate direction of research deals with visual interpretation of ViT attention heads. \cite{yeh2023attentionviz} explore how ViT attention heads display semantic patterns by visualizing the interactions between query and key embeddings. This approach allows one to see semantic patterns, providing insight into how different attention heads contribute to the understanding and processing of model input. But \cite{li2023does} also discuss the semantics of attention heads in ViTs by analyzing how different attention heads capture and represent various semantic patterns in images. This analysis helps identify which heads are more important and how their learned patterns relate to image content, enhancing the interpretability of ViTs.

\textbf{AI-generated image (AIGI) detection}.
%
Many promising approaches already exist for AIGI detection. 
\cite{ojha2023towards} use the ViT encoder embedding space from CLIP.
%
SubsetGAN \citep{cintas2022pattern} is based on the divergence of the discriminator's activation distributions for real and fake images. \cite{zhu2023gendet} generalize to unseen generators with an adversarial teacher-student discrepancy-aware framework, which focuses on creating larger output discrepancies between real and fake images during detection. 
Recent research concentrates on GAN-generated and especially diffusion-generated images. A cross-attention architecture for generic AIGI detection was developed by \citet{xi2023aigenerated}, \citet{porcile2024finding} concentrated on fake face detection with a CNN based on EfficientNet, and \citet{ha2024organic} studied artworks. \citet{moskowitz2024detecting} use pretrained CLIP embeddings and fine-tune the model and heads for AIGI detection, while \citet{bi2023detecting} do almost pure representation learning, finding latent representations characteristic for real images. \citet{cozzolino2023raising} also use CLIP embeddings, extracting captions from real images, generating the corresponding fake images by these captions, and training a robust SVM classifier on their embeddings. \cite{ricker2022towards} adapt GAN-generated detection methods to recognize diffusion-based images by retraining to detect specific artifact patterns and frequency characteristics unique to diffusions. \citet{ma2023exposing} exploiting the characteristic noise profiles of diffusion-based generators, while \cite{wang2023dire} detect diffusion-generated images by measuring the reconstruction error between the image and its reconstruction by a pretrained diffusion model.
Some works concentrate on texture details: PatchCraft~\citep{zhong2023rich} detects AIGI with inter-pixel correlation contrasts between rich and poor texture regions, while \citet{chen2024single} analyzes local image patches with simple textures for camera noise patterns. \citet{epstein2023online} study the generalization of AIGI detection methods to unseen models (presumably newly released generative models), with encouraging results: AIGI detection does generalize to new models, especially as the training set of existing models becomes richer; they also consider AI inpainting detection in an image. \citet{vahdati2024deepfake} take the next step from images to videos, noting that existing AIGI detectors do not generalize to videos. 
%
Our work also deals with robust AIGI detection, aiming to develop algorithms transferable to new generators without any adaptation. Previous works show that simple linear classifiers on top of pretrained semantic image embeddings (e.g. CLIP) are still a strong baseline, but training data selection is crucial \citep{wang2020cnn, epstein2023online}.

\section{Data}\label{sec:data}

One of the main challenges in fake image detection is constructing a dataset and choosing evaluation metrics.
%
There exist many different image generation models, including GANs, DALL-E, Stable Diffusion, and more, and new models keep appearing. If a training set for AIGI detection is based on some of these models, the model may overfit to specific generative models. The common practice here is to train a model on images produced by only one of these generators and then test it on fake images from other generators. This helps ensure that the model can generalize across generators, but the drawback is that the generative model chosen for the training set might be not the most effective to construct a universal training set.
%
Another issue is related to biases in the data. The final goal here is to create a robust solution for AIGI classification, so we want the training data to be balanced between two classes and unbiased, i.e., we want the only difference between fake and real images to be their ``fakeness''. Moreover, datasets of both fake and real images should be rich enough to cover different kinds of images, as we want the solution to work ``in the wild''. It is hard to come up with a procedure of creating such a balanced and unbiased dataset.
Due to these challenges, there is still no common benchmark for the task of fake image detection. Many works use manually created datasets, which are often biased and do not follow the distribution of real world images. Lately, due to rapid progress in image generation models, richer and more diverse datasets were published, which has made it easier to test the generalization abilities of classification models. These include the dataset by \cite{ojha2023towards} (13 CNN-based GANs, public), its extended version (with 8 diffusion models added, partially available), ArtiFact (over 2.5M images in total from 25 image generation models, including 7 diffusion-based ones, public) \citep{rahman2023artifact}, and the dataset by \cite{cozzolino2023raising} (18 models, both GAN-based and diffusion-based, not available).

Thus, we create a new dataset for the task of fake image detection that includes a diverse set of models and image semantics and can account for bias in the data. The dataset will be made publicly available upon acceptance, and we encourage the research community to contribute to it to make it richer and up-to-date with new generative models. 
We downloaded (image, text) pairs from the LAION-Aestethics dataset with the images aesthetics score $\geq$ 4.5 \citep{schuhmann2022laionb}. LAION is an open dataset
with a very diverse set of images; we filter by the aesthetics score to obtain images of at least moderate visual quality and thus make 
the distributions of real and generated images closer, as modern text-to-image models generally produce images of high visual quality. Based on the text prompts, we generate two images for every prompt using four modern diffusion-based text-to-image models:
DALL-E mini \citep{Dayma_DALLE_Mini_2021}, GLIDE with and without CLIP guidance \citep{DBLP:conf/icml/NicholDRSMMSC22}, Stable Diffusion v1.4 with 50 or 200 generation steps \citep{rombach2022high}, and Kandinsky-v2  with 20 or 100 generation steps \citep{razzhigaev2023kandinsky}. The resulting dataset contains 1001 image for each GAN-based model and 998 images for each diffusion model.

The goal of this procedure is to ensure that real and fake images have similar distributions in the dataset; we believe that the resulting set of images is diverse enough to reflect the distribution of real-world images. We suggest that this dataset can be successfully used for several different purposes:
\begin{inparaenum}[(1)]
    \item to train and test AIGI detection models in different settings, testing their robustness across generators;
    \item to see how an AIGI detection model's performance is affected by the generative model's architecture or the number of steps used in generation;
    \item to ensure the stability of an AIGI detection model by testing on two subsets of generated images produced from the same prompts;
    \item to compare the performance of different text-to-image models.
\end{inparaenum}
In our experiments we used the dataset with GAN-generated fake/real images by \cite{ojha2023towards} and our dataset produced with diffusion-based models. Our dataset creation procedure has similarities with the \emph{GenImage} dataset \citep{zhu2023genimage}, but we use more diverse and more detailed descriptions and generate several images per description; besides, \emph{GenImage} had not been released at the time of our research.

\section{Methods}\label{sec:methods}


\textbf{Interpreting CLIP-based fake image detection}.
CLIP image embeddings provide good features for distinguishing between real and generated images; on them, even simple models such as logistic regression or SVM yield state of the art results in AIGI detection, generalizing well across generative models \citep{ojha2023towards,cozzolino2023raising}. But their robustness highly depends on the training data generators, and there is plenty of room for improvement even for the best approaches. Interpretation of AIGI detection methods would help learn more about their mechanisms, identify failure cases, detect biases in data, and improve performance.

Our AIGI detection method is illustrated in Fig.~\ref{fig:method}a. It is based on the fact that CLIP embeddings of images and text can be compared via cosine similarity
$\cos(\x, \y) = {(\x^\top\y)} / {(\|\x\|\cdot \|\y\|)}.$
 We train a logistic regression model (LR) without the bias term for real/generated images classification using CLIP image embeddings as data. LR fits a weight vector of the same dimension as the embeddings, and classification probabilities monotonically depend on the dot product $\w^\top\x$: 
$\LR(\e^{\img}, \w) = \sigma\left( \w^\top\e^{\img}
\right),$
where $\e^{\img}$ is the image embedding and $\sigma$ is the logistic sigmoid $\sigma(a) = 1 / (1 + e^{-a})$. Thus, the normalized weight vector $\w$ can be interpreted as a vector in the CLIP embedding space, and the output of logistic regression for an image is based on cosine similarity between the learned weight vector and image embedding. Therefore, we can try to interpret the meaning of $\w$ by finding text tokens that have the most similar CLIP embeddings to the weight vector in terms of the cosine distance:
$\tok = \arg\,\max_{i}\left(\w^\top\CLIP(t_i)\right)$,
where $t_i$ is the $i$-th vocabulary token. Below, we train LR models on fake images produced by different generative models and show that their weight vectors can be interpreted using our method. We report a positive correlation between the interpretability of a model's weight vector $\w$ and low generalization of this model to images generated by other generative models.
We also show that LR weight vectors are close to the residual vectors of generative models in terms of the cosine distance. We define the residual vector of a generative model as
$\res_g = \mathrm{Norm}\left[\frac{1}{M}\sum\nolimits_{m=1}^M \e^g_m \right] - \mathrm{Norm}\left[ \frac{1}{N}\sum\nolimits_{n=1}^N \e^r_n \right],$
where $N$ and $M$ are the numbers of real and fake images in the training set, 
$\e^r_n$ and $\e^g_m$ are CLIP embeddings of the $n$-th real image and the $m$-th image produced by generative model $g$,
and $\mathrm{Norm}$ denotes normalization. In a way, the residual vector transforms an embedding of a real image into the embedding of a fake image. Thus, we model the probability of an image to be fake based on the residual vectors as
$f_{\res_g}(e) = ({\e^\top \res_g}) / ({\|\e\|\cdot \|\res_g\|}),$
where $\e$ is the image's CLIP embedding.
By choosing a threshold, we get a binary classifier for fake/real images; this can be done by fitting the LR model on samples of the form $\{ \e^g_m \cdot \res_g \}^N_{m=1}$, where $\e^g_m$ is the CLIP embedding of the $m$-th image produced by generator $g$. This model is easy to train and highly interpretable. 


\def\dsearch{D_{\mathrm{search}}}
\def\deval{D_{\mathrm{eval}}}

\textbf{Removing features to improve robustness}.
Although CLIP-based fake image detectors show excellent results in-domain, their performance drops on generators not present in the training set. One effective method to improve the robustness of a machine learning model is to remove unnecessary features. If we can find features in the embeddings capturing generator-specific information, their removal could improve the classifier's generalization ability.
To find such features, we apply an iterative greedy search algorithm on a separate subset of data. In our dataset (see Section~\ref{sec:data}), we select a pair of generators and find features that yield the best cross-generator transfer results (Fig.~\ref{fig:method}b, top). 

Formally, for a set of $I$ domains $D_i$ (a domain corresponds to a generator model) 
we divide the data into $\dsearch = \{D_1, D_2\}$ with two domains and $\deval = \{D_i| i\neq 1, 2\}$.
On every step, we start with a feature vector of dimension $d_s$ and find the component whose removal maximises the out-of-domain score on $\dsearch$ measured from $D_1$ to $D_2$, thus training $d_s$ separate classifiers. Then we drop this coordinate and perform another search step, ultimately constructing an ordered list of candidates for removal $L_{1\rightarrow 2}$ and the corresponding scores $S_{1\rightarrow 2}$. We perform the same procedure for the out-of-domain score from $D_2$ to $D_1$, getting lists of candidates $L_{2\to 1}$ and scores $S_{2\to 1}$. 
Then we choose $\alpha=\arg\max_\alpha S(\alpha)$ in both lists, remove all features before $\alpha$, and take the union of the remaining lists. Interestingly, in our experiments the score increases until over $90\%$ of the features are removed (see Fig.~\ref{fig:progan_sd} in the Appendix).   
%
After the subset of features is selected, we consider domains from $\deval$ for measuring the effectiveness of domain transfer in AIGI detection. For each domain $D_i\in \deval$, we train a classifier using fake data from $D_i$, and then test it on all other domains $D_j\in\deval$, reporting scores for each pair $D_i, D_j \in\deval$.  





\textbf{AIGI detection with attention head outputs}.
This approach is inspired by \cite{gandelsman2024interpreting} who show that every attention head's output in the CLIP model has a direct linear impact on the final embedding and map attention head outputs into the CLIP embedding space. This mapping allows to interpret the semantics of every head by collecting text descriptions and analysing the cosine similarity between CLIP text encoder embeddings of these descriptions and mapped attention head outputs. We consider the embedding of each head's output as a separate subset of features, and select the best \textit{heads} for cross-domain transfer. Formally, we begin with the representation of an image $I$:

\noindent
\begin{equation}
\label{eq:vit_decomposition}
M_{image}(I)=P\text{ViT}(I)=P[Z^0+\sum\nolimits_{i=1}^L\text{MSA}^l(Z^{l-1})+\sum\nolimits_{i=1}^L\text{MLP}^l(\hat{Z^{l}})],
\end{equation}

\noindent
where MSA corresponds to multi-head self-attention components, and MLP denotes multi-layer perceptron outputs. Each MSA term can be further decomposed into individual head impacts:

\noindent
\begin{equation}
\label{eq:heads}
\text{MSA}^l(Z^{l-1})=\sum\nolimits_{h=1}^H \sum\nolimits_{i=1}^N \alpha_i^{l,h} W_{VO} Z_i^{l-1} = \sum\nolimits_{h=1}^H E^{l,h}(I),
\end{equation}

\noindent
where $E^{l,h}(I)$ denotes head-wise image embeddings. To construct our head-based fake text detector, we select a subset of heads $\mathcal{H}$ with embeddings $\{E^{l_i, h_j} | (l_i, h_j)\in \mathcal{H}\}$  and learn a logistic regression classifier on top of the concatenation of these embeddings. After choosing the optimal subset of heads $\mathcal{H}^{\mathrm{best}}$, we may obtain an \textit{explanation} of the classifier from semantic interpretations of the projections $\{PE^{l_i, h_j} | (l_i, h_j)\in \mathcal{H}^{\mathrm{best}}\}$ in the joint text-image space  (Fig.~\ref{fig:method}b, bottom).
The optimal subset of heads is chosen with the following algorithm. We first fix two models: one for training and one for validation. For each attention head, we train Logistic Regression on its outputs on images from the training set (produced by the model chosen for training) and validate on images from the validation set. We then find 3 top-performing attention heads in this setting. After that, we fix these attention heads and train LR models on data from every model as training one and test on data from every other model. For training hyperparameter search, the same validation set is used as for choosing attention heads. Model used for validation is excluded from training and testing models. We conduct several such experiments using different training-validation pairs to prove robustness.

\section{Experimental evaluation}\label{sec:eval}

In this section, we describe the results of our experimental evaluation; a detailed description of the experimental setup, preprocessing, and hyperparameters is given in the Appendix.


\textbf{Baseline cross-generator transfer.} For a set of images generated by models $G_1, G_2,\ldots, G_I$, we train logistic regression (LR) on images generated by $G_i$ and then test it on images generated by $G_j$, $j\neq i$. For GAN-based models, we use train/test splits provided by \cite{ojha2023towards}; for diffusion-based models, we use a 7:3 train:test split; train and test sets for different models are based on the same real images to avoid data leaks across models. The resulting accuracies are reported in Fig.~\ref{fig:clip_embeddings} in the Appendix. 
%
We trained LR both with and without the bias term, with similar performance.
%
We also compute $\res_g$ and train LR models $f_{\res_g}$ for each generator (Fig.~\ref{fig:clip_residuals} in Appendix).
For most models, generalization accuracy is similar to that of LR trained on full CLIP image embeddings, and the residual vector is very close to the weight vector of LR trained on data from this model.
%
%
We observe that, first, generative models differ in terms of generalizability. LR trained on SD-1.4 transfers much better than trained on DALL-E or BigGAN. 
DALL-E and GLIDE are ``in-between'' here, perhaps
because they are diffusion-based but their image quality is poor compared to the latest diffusion models. In general, the transfer from diffusion-based models to GANs and back is the most difficult here, so we report average accuracy of transfer between these two groups (Table~\ref{tbl:results}, right).

\textbf{Interpretation of the classifier weights.} Table~\ref{tbl:lrtokens} shows the interpretation of LR weights
according to Section~\ref{sec:methods}. 
We also find text tokens whose embeddings are nearest to $\res$ of each model (Table~\ref{tbl:lrtokens_residuals} in the Appendix), and note that the semantics of both methods are very similar, so below we discuss the interpretation of LR weights.
We can see that words with nearest embeddings indeed often express certain properties of generated images, e.g., ``blurry'' and ``blur'' for GLIDE, 
``uv'' for the Kandinsky model, and ``gouache'' for SD-1.4. Words such as ``bild'' or ``vscocam'' may also suggest certain salient characteristics. For some GAN-based models, nearest embeddings include words such as ``deeplearning'',  ``generative'', and ``neural'', especially for residuals, which suggests that GAN-generated images with such captions were present in the training set of CLIP. This also might explain why CLIP embeddings turn out to be good features for GAN-produced image detection.

Second, we can spot biases in the data. E.g., the word with the nearest embedding to LR weights and residual of StarGAN is ``schwarzenegger'', likely because StarGAN is generating human faces, so the training data for AIGI detection consists entirely of faces. 
This confirms that our approach does capture image properties characterizing generative models. 
We also find nearest neighbours for LR weights among phrases from \citep{gandelsman2024interpreting}, intentionally generated to describe image properties. Table~\ref{tab:phrases} shows that GANs tend to generate detailed fantastic illustrations, while diffusion-based models produce photorealistic scenes, which is supported by illustrations shown in Table~\ref{tbl:lr-images}.
%
Interestingly, we find a negative correlation between a model's generalizability and its ``interpretability'', i.e., max similarity to text tokens. The best generator in terms of generalization power is SD-1.4; it also has the lowest similarity between its LR weights and either residual or token embeddings, with the nearest neighbor having cosine similarity $0.08$ (for other models it is $\approx 0.12$-$0.14$). LR on the residuals of SD-1.4 performs about $10\%$ worse, and its max cosine similarity to token embeddings rises to $0.15$. Thus, we find that the weight vector of a generalizable LR model should not be too similar to any vector in the CLIP embedding space with a distinct meaning.

\def\aall{A_{\mathrm{all}}}
\def\adiff{A_{\mathrm{diff}}}
\def\agan{A_{\mathrm{GAN}}}
\def\agd{A_{\mathrm{GAN}\to\mathrm{diff}}}
\def\adg{A_{\mathrm{diff}\to\mathrm{GAN}}}

Table~\ref{tbl:results} summarizes the generalization performance of LR trained on CLIP embeddings and residual features, showing transfer results between GAN-based and diffusion-based generators. 
%
We compute 5 metrics: 
the average value of the entire accuracy matrix for cross-domain transfer, 
average over domains corresponding to GAN-based architectures, 
for domains corresponding to diffusion-based models, 
average cross-domain generalizability from GAN-based to diffision-based models and back.
%

%

%
\textbf{Feature selection results.} For experiments with greedy feature search, we chose 
five random pairs as $\dsearch$ and 
evaluated feature selection
on the rest of the models, training LR on each model and evaluating it on other models from $\deval$; then we report the average accuracy of these classifiers. We test our algorithms on two CLIP image encoders, ViT-B-16 (CLIP-base) and ViT-L-14 (CLIP-large), reducing the feature space by $5$-$10$x. Table~\ref{tbl:results} shows that results improve by $1$-$3$\% for any choice of $\dsearch$ and both encoders. The best average transfer between groups of models
is achieved on GAN detection with a diffusion-trained classifier by CLIP-large (+$6$\%).

\textbf{Head selection and interpretation}. The two domains in $\dsearch$ have different function for head selection, so Table~\ref{tbl:results} reports two numbers for each $\dsearch$. On average, this method works best for the transfer from diffusion to GANs ($+7.6$\%) on CLIP-base. We find significant improvements for almost all pairs, with the best $+6$\% obtained with StyleGAN/GLIDE-CLIP and GauGAN/GLIDE-base as train/val pairs.
%
Next we find semantic interpretations of the heads chosen on the best $\dsearch=\{\text{StyleGAN}, \text{GLIDE-CLIP}\}$. Table~\ref{tbl:headsnn} shows that head $(4,6)$ indeed has semantics very close to typical residuals from Table~\ref{tab:phrases}, such as ``\emph{Detailed illustration of smth. futuristic}'' or even just ``\emph{Generated image}''; head $(6,0)$ is responsible for bright, expressive and even ``\textit{psychodelic}'' colors; $(5,5)$ detects something ``\textit{whimsical}'', ``\textit{twilight}'', or ``\textit{lunar}'', which is opposite to practical objects such as ``\textit{plastic}'' and ``\textit{garbage truck}'' (its farthest tokens). In general, it corresponds to our intuitive understanding of the typical differences between real and generated images.

\begin{table}\centering\setlength{\tabcolsep}{2pt}\footnotesize
\begin{tabular}{cX{.39\linewidth}l|cX{.39\linewidth}l}\toprule
 \multirow{5}{*}{\rotatebox{90}{\textbf{GLIDE-base}}} & Fast-paced race car blur & 0.181 
&  \multirow{5}{*}{\rotatebox{90}{\scriptsize\textbf{Kandinsky-100}}}
& Det. ill. futuristic energy generator & 0.242 \\
& Det. ill. futuristic brain-computer interface & 0.146 
& & Det. ill. futuristic brain-computer interface & 0.239 \\
& Blurred abstraction & 0.138 
& & Det. ill. futuristic biotechnology & 0.212 \\
& Det. ill. alien world & 0.126 
& & Det. ill. alien world & 0.205 \\
& Illustration of an alien landscape & 0.126 
& & Det. ill. futuristic medical breakthrough & 0.202 \\\midrule
  \multirow{5}{*}{\rotatebox{90}{\textbf{SD-1.4-200}}} & 
Vibrant city alley & 0.128
& \multirow{5}{*}{\rotatebox{90}{\scriptsize\textbf{DALL-E-1059}}}
& Serene beach sunset & 0.126 \\
& Det. ill. alien world & 0.116 
& & Surreal photo manipulation & 0.123 \\
& Photo featuring a vibrant street graffiti & 0.109
& & Photo with grainy, old film effect & 0.118 \\
& Det. ill. futuristic brain-computer interface & 0.092
& & Photo with vintage film grain effect & 0.115 \\
& Photo featuring a vibrant urban graffiti & 0.090
& & tranquil beach sunset & 0.111 \\\midrule
  \multirow{5}{*}{\rotatebox{90}{\textbf{GauGAN}}} & 
Image with a double exposure effect & 0.245 
& \multirow{5}{*}{\rotatebox{90}{\textbf{StyleGAN 2}}}
& Photo with sepia-toned vintage style & 0.096 \\
& Ethereal double exposure photography & 0.242
& & Photograph taken in a rustic barn & 0.092 \\
& Double exposure effect & 0.241
& & Deserted coastal pier & 0.091 \\
& Image with double exposure effect & 0.240
& & Vintage sepia tones & 0.091 \\
& Impressionist-style digital painting & 0.237
& & Photo with soft, dreamy tones & 0.085 \\
\midrule
  \multirow{5}{*}{\rotatebox{90}{\textbf{ProGAN}}} 
& Impressionist-style digital painting & 0.191
& \multirow{5}{*}{\rotatebox{90}{\textbf{BigGAN}}}
& Generated image & 0.214 \\
& {\scriptsize Photograph with the artistic style of double exposure} & 0.189
& & Generated photo & 0.165 \\
& Det. ill. futuristic brain-computer interface & 0.188
& & GAN generated image & 0.162 \\
& Det. ill. futuristic virtual realm & 0.179
& & Impressionist-style digital painting & 0.132 \\
& Surreal digital collage & 0.178
& & Det. ill. futuristic brain-computer interface & 0.125 \\
\bottomrule
\end{tabular}
\caption{Nearest neighbor phrases for image generators. ``Det. ill.'' -- ``Detailed illustration of a''.}
\label{tab:phrases}
\end{table}

\begin{table}[!t]\centering\setlength{\tabcolsep}{3pt}\footnotesize
\begin{tabular}{l|X{.39\linewidth}l|X{.39\linewidth}l}\toprule
 & \multicolumn{2}{c|}{\textbf{Nearest to CLIP heads}} & \multicolumn{2}{c}{\textbf{Farthest from CLIP heads}} \\\midrule
\multirow{5}{*}{\rotatebox{90}{\textbf{Head (4, 6)}}}
& Det. ill. futuristic biotechnology & 0.070 
& Picture of fast food & -0.115 \\
& Generated image & 0.069 
& An image of three subjects & -0.109 \\
& Det. ill. futuristic medical breakthrough & 0.068 
& A zoomed in photo & -0.103 \\
& Det. ill. futuristic medical technology & 0.067 
& Vibrant watercolor painting & -0.097 \\
& Urban labyrinth & 0.062 
& Ethereal double exposure photography & -0.096 \\\midrule
\multirow{5}{*}{\rotatebox{90}{\textbf{Head (5, 5)}}} 
& Whimsical composition & 0.076 
& An image of two subjects & -0.059 \\
& An image of a Chiropractor & 0.076 
& Image of a garbage truck & -0.052 \\
& Photo with cool, twilight tones & 0.075 
& A photograph of a medium-size object & -0.052 \\
& Cultural mosaic & 0.074 
& Marbleized design & -0.050 \\
& Image with a lunar eclipse & 0.072 
& Close-up of a textured plastic & -0.049 \\\midrule
\multirow{5}{*}{\rotatebox{90}{\textbf{Head (6, 0)}}} 
& Colorful expressions & 0.134 
& Surreal photo manipulation & -0.074 \\
& colorful celebration & 0.118 
& {\scriptsize Image with a futuristic augmented reality scene} & -0.067 \\
& Psychedelic color swirls & 0.118 
& Futuristic drone technology & -0.064 \\
& A fern & 0.113 
& advanced drone technology & -0.058 \\
& colorful ceremony & 0.112 
& Image with sand and dust & -0.057 \\\bottomrule
\end{tabular}

\caption{Interpretation of the best performing heads of CLIP-ViT model.}\label{tbl:headsnn}
\end{table}

\begin{table*}[!t]\centering\setlength{\tabcolsep}{2.2pt}\footnotesize
\begin{tabular}{rl|rl|rl|rl|rl|rl } 
 \toprule
  \multicolumn{2}{c}{$ $} & \multicolumn{2}{c|}{SD-1.4-200} & \multicolumn{2}{c|}{DALL-E-mini} & \multicolumn{2}{c|}{GLIDE-base} & \multicolumn{2}{c|}{GLIDE-CLIP} & \multicolumn{2}{c}{Kandinsky-100} 
  \\\midrule
\multicolumn{3}{r}{gouache} & 0.08 & instaweather & 0.15 & blurred & 0.13 & blurred & 0.12 & uv & 0.12 \\ 
\multicolumn{3}{r}{방탄소년} & 0.08 & 태 & 0.11 & blurry & 0.12 & blurry & 0.11 & fluor & 0.11 \\
\multicolumn{3}{r}{bild} & 0.08 & webcamtoy & 0.10 & blur & 0.08 & 방탄소년 & 0.08 & renders & 0.10 \\
\multicolumn{3}{r}{instaweatherpro} & 0.08 & gouache & 0.10 & 방탄소년 & 0.08 & octane & 0.08 & build & 0.10 \\
\multicolumn{3}{r}{vscocam} & 0.07 & piccollage & 0.10 & octane & 0.08 & weil & 0.08 & busan & 0.09 
\\\midrule
%
\multicolumn{2}{c|}{GauGAN} & \multicolumn{2}{c|}{CycleGAN} & \multicolumn{2}{c|}{StyleGAN} & \multicolumn{2}{c|}{ProGAN} & \multicolumn{2}{c|}{BigGAN} & \multicolumn{2}{c}{StarGAN} 
\\\midrule
 manatee & 0.15 & manatee & 0.14 & piccollage & 0.12 & piccollage & 0.14 & instaweather & 0.19 &  {\scriptsize schwarzenegger} & 0.14 \\ 
 digital & 0.13 & {\scriptsize deeplearning} & 0.13 & stamatic & 0.09 & manatee & 0.12 & stamatic & 0.17 & transpa & 0.14 \\
 bungal & 0.12 & generative & 0.11 & instaweather & 0.09 &  {\scriptsize instaweather} & 0.11 & png & 0.17 & oprah & 0.13 \\
simulated & 0.12 & scanned & 0.10 & bharti & 0.06  & generative & 0.10 & piccollage & 0.16 & hani & 0.13 \\
 scrapped & 0.12 & effects & 0.10 & {\scriptsize instaweatherpro} & 0.06  & orthodon & 0.10 &  {\scriptsize instaweatherpro} & 0.15 & zlatan & 0.13 \\
\bottomrule
\end{tabular}
\caption{Interpretation of LR weights $\w$ trained on fake/real data; each column shows 5 words whose CLIP embeddings are nearest to $\w$ and the corresponding cosine similarities.}\label{tbl:lrtokens}
\end{table*}

\begin{table}[!t]\centering\setlength{\tabcolsep}{3pt}\small
\begin{minipage}{.62\linewidth}
\begin{tabular}{rr|cccccc}
\toprule
\multicolumn{2}{c}{\textbf{Feature selection}} & \multicolumn{2}{c}{\textbf{CLIP-base}} & \multicolumn{2}{c}{\textbf{CLIP-large}} \\
\textbf{Validation} & \textbf{Training} & \textbf{Heads} & \textbf{Comp.} & \textbf{Heads} & \textbf{Comp.} \\\midrule
\multicolumn{2}{c|}{Baseline} & 0.722 & 0.725 & 0.783 & 0.783 \\\midrule
GLIDE-CLIP & StyleGAN & 0.784 & \multirow{2}{*}{0.737} & 0.754 & \multirow{2}{*}{0.787} \\
StyleGAN & GLIDE-CLIP & 0.764 & & 0.748 \\\midrule
CycleGAN & Kandinsky-20 & 0.746 & \multirow{2}{*}{\textbf{0.758}} & 0.779 & \multirow{2}{*}{\textbf{0.812}}\\
Kandinsky-20 & CycleGAN & 0.776 & & \textbf{0.806} \\\midrule
DALL-E-mini & ProGAN & 0.742 & \multirow{2}{*}{0.740} & - & \multirow{2}{*}{-}\\
ProGAN & DALL-E-mini & 0.759 & & -\\\midrule
GauGAN & GLIDE-base & 0.671 & \multirow{2}{*}{0.747} & 0.716 & \multirow{2}{*}{0.794} \\
GLIDE-base & GauGAN & \textbf{0.785} & & 0.672 & \\\midrule
StarGAN & Kandinsky-100 & 0.728 & \multirow{2}{*}{0.745} & - & \multirow{2}{*}{-} \\
Kandinsky-100 & StarGAN & 0.758 & & -
\\\bottomrule
\end{tabular}
\end{minipage}~~~~\begin{minipage}{.32\linewidth}
\includegraphics[width=\linewidth]{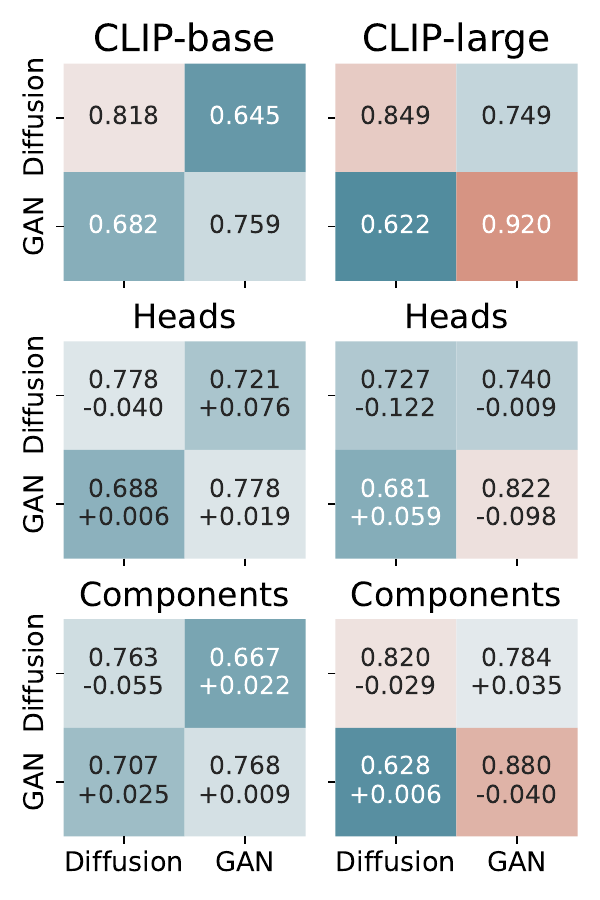}
\end{minipage}

\caption{AIGI detection results. Left: average performance of LR classifiers. Right:  transfer accuracy between diffusion-based and GAN-based generators averaged over feature selection pairs. }\label{tbl:results}
\end{table}

\textbf{Qualitative visual analysis}
In this section, we analyse the images where LR shows the best and worst performance. We use LR trained on images from SD-1.4-50 and ProGAN generators. The results are presented in Table~\ref{tbl:lr-images}.
%
    We find that, first, the image resolution is important for AIGI detection:
    images correctly and confidently classified as real are mostly low-res while real images confidently misclassified by LR are high-res.
    This may be caused by the fact that images produced by generative models are mostly high-resolution, with lots of details.
    Interestingly, our semantic interpretation also proposes ``Detailed'' as an important characteristic of images generated by several models, including SD (Table~\ref{tab:phrases}). Data augmentation such as lowering the resolution of some fake images for LR training may further improve AIGI detection.  
    Second, there are many images with repeating patterns among those generated by SD-1.4 and misclassified by LR. This possibly shows a bias of some generative models towards generating images with repetitive patterns. 
    Images correctly classified as AI-generated
    show clear artifacts such as over-exposition, fuzzy or unnatural shapes etc., while fake images misclassified as reals are clear and have realistic colors. This applies to both SD-1.4 and ProGAN models and is supported by the fact that Table~\ref{tbl:lrtokens} often speaks of popular photo filters and editors: \textit{instaweather}, \textit{vscocam}, \textit{webcamtoy} \textit{gouache}, \textit{blurred}. 
    Moreover, ProGAN semantic interpretation contains phrases clearly describing this difference: ``\textit{Surreal digital collage}'' and ``\textit{Photograph with the artistic style of double exposure}'' both have very high similarity score 0.24.
    

\def\picwid{20mm}
\def\pichei{20mm}
\newcommand\mypic[1]{\includegraphics[width=\picwid, height=\pichei]{figures/lr_images/#1.png}}

 \begin{table*}[!t]
 \centering\setlength{\tabcolsep}{2.5pt}
      \begin{tabular}{ c | ccc | ccc}
      \toprule
 & \multicolumn{3}{c|}{\textbf{Stable Diffusion-1.4-50}} & \multicolumn{3}{c}{\textbf{ProGAN}} \\\midrule
{\rotatebox{90}{\textbf{Real$\to$Real}}} & 
\mypic{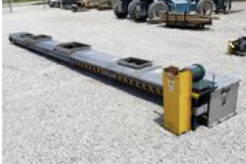} & \mypic{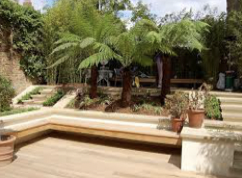} & \mypic{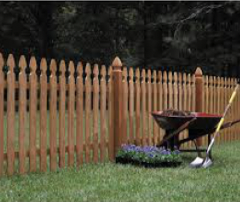} & 
\mypic{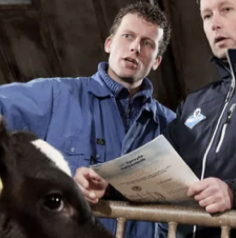} & \mypic{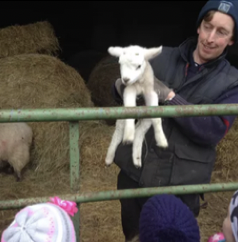} & \mypic{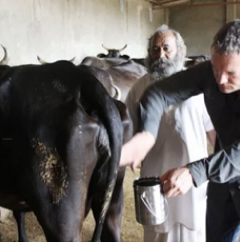} \\
& {\footnotesize $180\times 180$} & {\footnotesize $259\times 194$} & {\footnotesize $255\times 255$}& & & \\\midrule
{\rotatebox{90}{\textbf{Real$\to$Fake}}} & 
\mypic{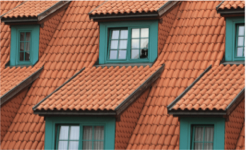} & \mypic{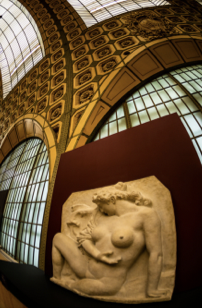} & \mypic{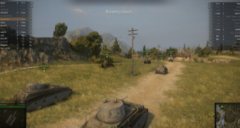} & 
\mypic{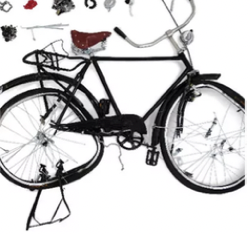} & \mypic{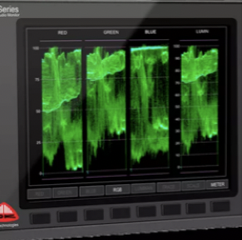} & \mypic{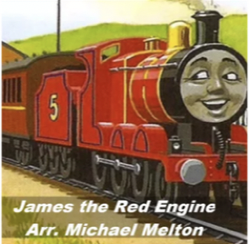} \\
& {\footnotesize $1080\times 660$} & {\footnotesize $501\times 750$} & {\footnotesize $1920\times 1080$} & & & \\\midrule
\rotatebox{90}{\textbf{Fake$\to$Fake}} & 
\mypic{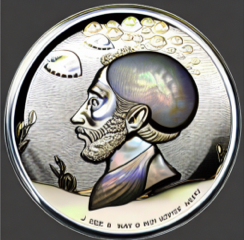} & \mypic{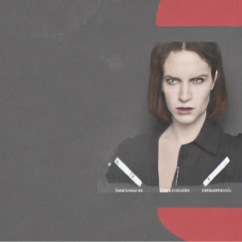} & \mypic{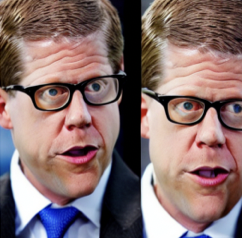} & 
\mypic{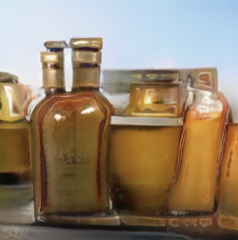} & \mypic{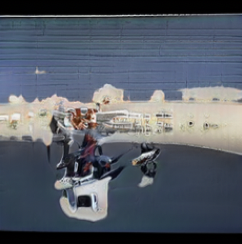} & \mypic{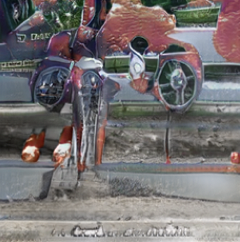} \\\midrule
\rotatebox{90}{\textbf{Fake$\to$Real}} & 
\mypic{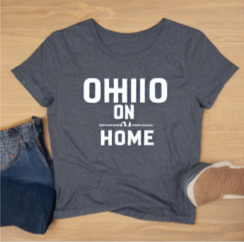} & \mypic{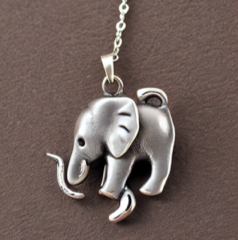} & \mypic{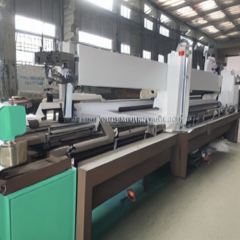} & 
\mypic{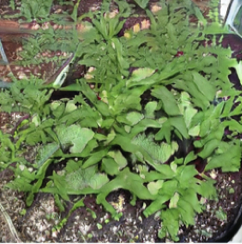} & \mypic{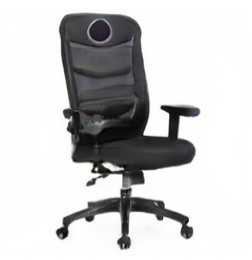} & \mypic{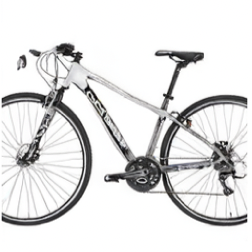} \\\bottomrule
\end{tabular}

\caption{Images from SD-1.4-50 and ProGAN test sets where LR shows the best and worst performance; e.g., ``Real$\to$Fake'' are real images most confidently classified as fake. All SD images are $512\times 512$ and ProGAN images are $256\times 256$ unless specified otherwise.}\label{tbl:lr-images}
\end{table*}


\section{Limitations and broader impacts}\label{sec:limits}

The increasing amount and quality of AI-generated content have raised difficult challenges and safety concerns, which this work aims to alleviate. We address the problem of robust detection for AI-generated images, using feature selection methods to reduce the dimensionality without quality degradation. However, our methods still require significant computations, being based on modern pretrained image embeddings. Moreover, in this work we do not cover the risks of false accusations of faithful content creators, which is important to consider in AI safety scenarios. On the other hand, we propose interpretation methods for AIGI detectors, which may help discover and mitigate biases, but specific ways to mitigate them require additional study.

\section{Conclusion}\label{sec:concl}

Simple linear classifiers on pretrained text and image embeddings are known to detect artificial content with high accuracy, but this approach is usually not reliable enough for real-life usage because these classifiers struggle with transfer across semantic domains or generator models.
Previous experiments with cross-domain and cross-generator transfer show that some out-of-domain data can radically change the performance of supervised detectors, in some cases dropping to random guessing values. 
%
In this work, we take a step towards overcoming this drawback,
improving the robustness of the classifier (up to 
9 \% in absolute accuracy in some cases) while actually simplifying the baseline algorithms. First, we use residual vector projections as a one-feature classifier instead of training LR or SVM on CLIP embeddings; we show that this usually does not hinder classification quality but allows for a semantic description of the difference between real and fake images, discovering biases and incompleteness in the training set (via text tokens such as ``blur'' or ``instaweather''). 
In fact, our recipe is to construct the training set so that there are no understandable tokens close to the residual; this kind of data leads to more robust classifiers.
The next surprisingly efficient idea is to reduce the dimension by removing components from pretrained embeddings.
We have shown that a 47-dimensional subset of CLIP embeddings outperforms the original 768-dim vector and, more importantly, allows non-trivial transfer from GAN-based to diffusion-based generators, improving from nearly random to 70\% accuracy for some train-test generator pairs. 
%
Our final ``simplification'' is achieved by Transformer head pruning. In general, our results provide further validation of the usefulness of joint text-image embeddings for AIGI detection; we hope that future research can build upon our results to achieve robust and practically useful AIGI detection.

\bibliographystyle{abbrvnat}
\bibliography{neurips_2024}

\appendix

\section{Detailed experimental results on the detection of generated images}

\subsection{Computational resources}
\label{sec:comp_resources}

For all of our experiments (including dataset generation), we used 3 servers with the following computational resources:
\begin{itemize}
    \item 1 V100 16Gb GPU + 32 CPUs (Intel(R) Xeon(R) Gold 6151), 126GB RAM
    \item 2 V100 16GB GPUs + 64 CPUs (Intel(R) Xeon(R) Gold 6151), 252GB RAM
    \item 1 A100 40GB GPU + 8 CPUs (Intel(R) Xeon(R) X3470), 502GB RAM
\end{itemize}
GPUs were used only for generating the dataset and for extracting the embeddings (including head embeddings) from the CLIP models; these calculations took less than 7 days in total. The rest of the calculations were performed on CPUs and took less than 7 days in total, as well.
The slowest CPU experiment (component selection for clip-vit-large-patch14 embeddings) took nearly 12 hours for one pair of models on the weakest server. 

The RAM usage never exceeded the RAM amount of the weakest server, and every used model could fit into one A100 40GB GPU, so 126 GB of RAM and 40 GB of VRAM are enough to reproduce each of our experiments.

\subsection{Full results for embeddings and residuals}

\begin{figure*}
    \centering\setlength{\tabcolsep}{0pt}
    \begin{tabular}{cc}
    \includegraphics[width=0.49\textwidth]{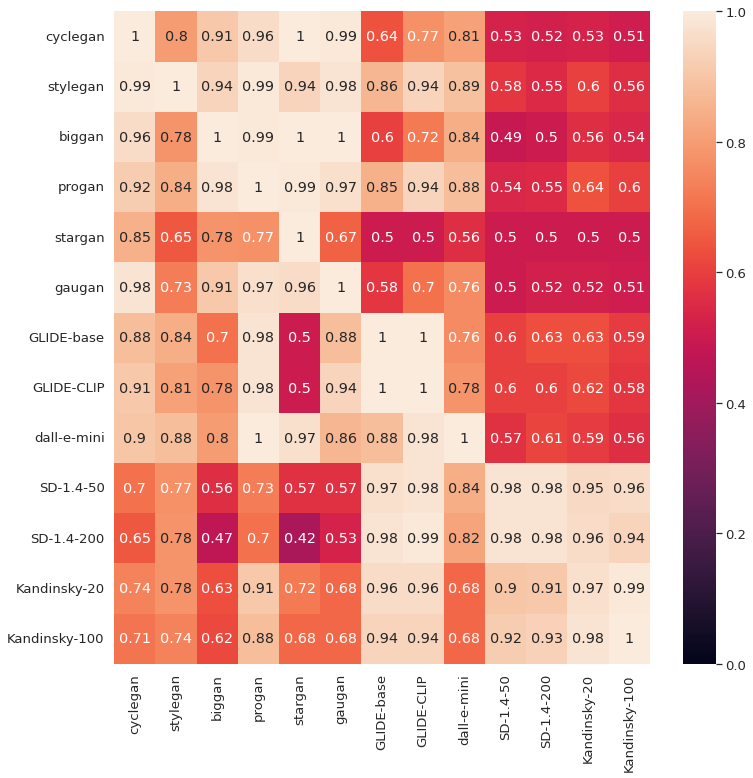} &
    \includegraphics[width=0.49\textwidth]{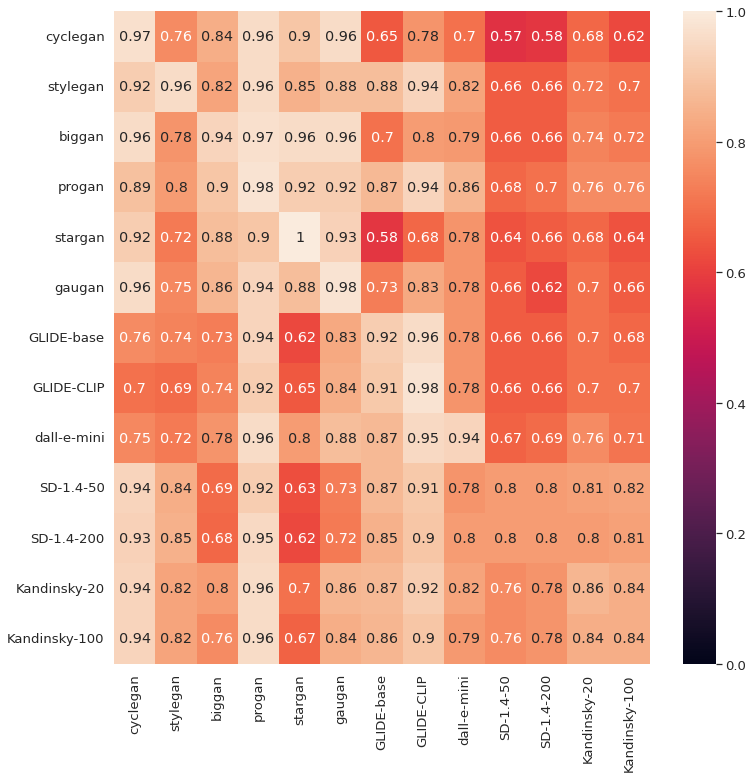}
    \end{tabular}
    \caption{Classification on CLIP embeddings: left~--- original embeddings (mean accuracy: 78.33\%; mean accuracy without SD-1.4.-200 and ProGAN: 77.74\%); right~--- embeddings where ``bad'' dimensions are removed (mean accuracy: 80.36\%; mean accuracy without SD-1.4.-200 and ProGAN: 79.69\%).}
    \label{fig:clip_embeddings}
\end{figure*}

\begin{figure*}
    \centering\setlength{\tabcolsep}{0pt}
    \begin{tabular}{cc}
    \includegraphics[width=0.49\textwidth]{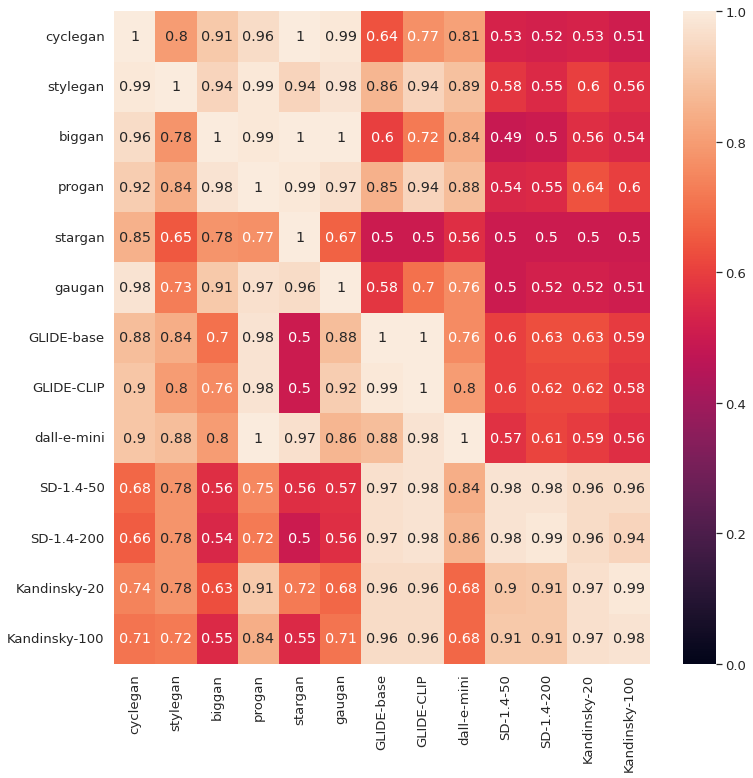} &
    \includegraphics[width=0.49\textwidth]{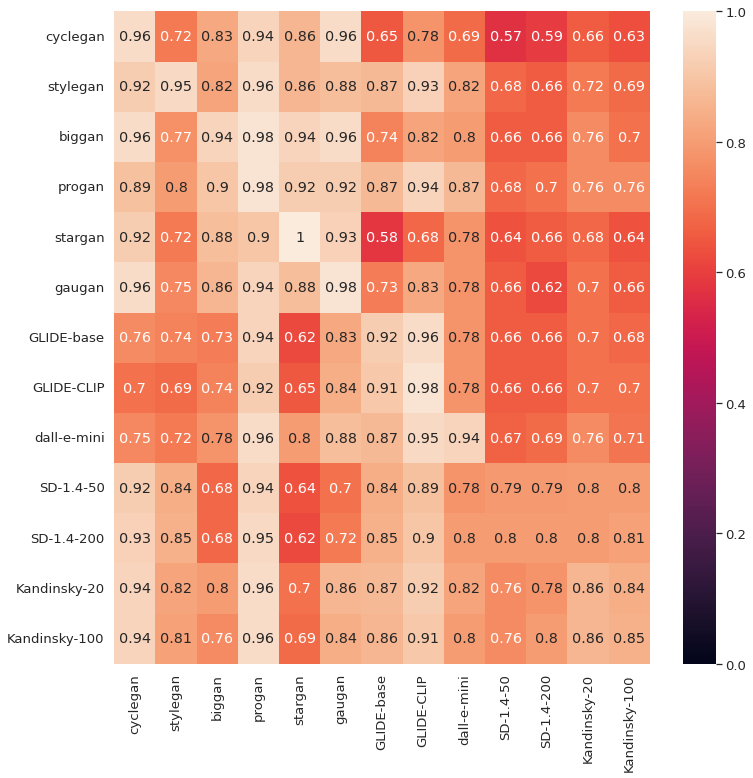}
    \end{tabular}
    \caption{Classification on CLIP embeddings with \texttt{fit\_intercept=False}: left~--- original embeddings (mean accuracy: 78.31\%); right~--- embeddings where ``bad'' dimensions are removed (mean accuracy: 80.31\%).}
    \label{fig:clip_embeddings_fifalse}
\end{figure*}

\begin{figure*}
    \centering\setlength{\tabcolsep}{0pt}
    \begin{tabular}{cc}
    \includegraphics[width=0.49\textwidth]{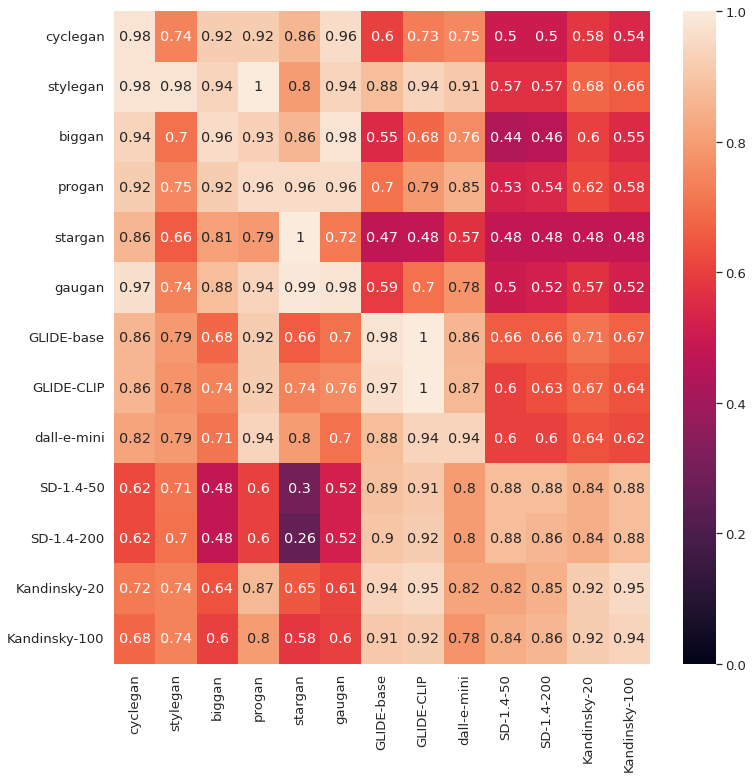} &
    \includegraphics[width=0.49\textwidth]{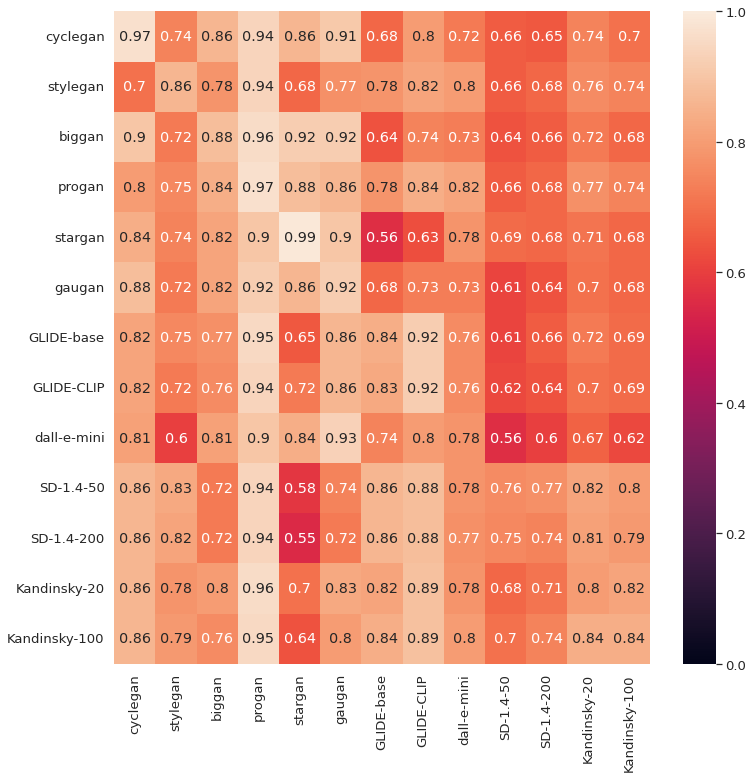}
    \end{tabular}
    \caption{Classification on CLIP residuals: left~--- original embeddings (mean accuracy: 75.39\%); right~--- embeddings where ``bad'' dimensions are removed (mean accuracy: 77.86\%).}
    \label{fig:clip_residuals}
\end{figure*}

In this section, we present full results for the experiments on transferring fake/real image classifiers across different image generators, using SD-1.4-200 and ProGAN as a typical pair for components search. Figure~\ref{fig:clip_embeddings} shows the results for CLIP embeddings with all components (dimensions) present (on the left) and with ``bad'' dimensions removed (on the right). In Figure~\ref{fig:clip_embeddings} and subsequent such plots, the vertical axis shows generators used for producing the ``fake'' part of the training set, and the horizontal axis shows the generators used for the test set. 

Figure~\ref{fig:clip_embeddings_fifalse} shows the same experiment but without the bias term in logistic regression (\texttt{fit\_intercept=False}); these results clearly show that removing the bias term has no negative effect on the results (no effect at all, in fact). Finally, Figure~\ref{fig:clip_residuals} shows the results of the same experiment for CLIP residuals.
%
Below we describe the methodology for constructing Figures~\ref{fig:clip_embeddings},~\ref{fig:clip_embeddings_fifalse}, and~\ref{fig:clip_residuals} in more detail.

Tables \ref{tab:phrases_residuals} and \ref{tbl:lrtokens_residuals} provide a comparison of semantic interpretation of embeddings and residuals. We see that the semantic descriptions for both methods are very similar, but the cosine similarity score for residuals is slightly larger for all the generators.

\begin{table*}[!t]\centering\setlength{\tabcolsep}{3pt}\footnotesize
\begin{tabular}{c|X{.35\linewidth}l|X{.45\linewidth}l}\toprule
  & \multicolumn{2}{c|}{\textbf{GLIDE-base}} & \multicolumn{2}{c}{\textbf{Kandinsky-100}} \\\midrule
\multirow{5}{*}{\rotatebox{90}{\textbf{CLIP embeddings}}} &
Fast-paced race car blur & 0.181 
& Detailed illustration of a futuristic energy generator & 0.242 \\
& Detailed illustration of a futuristic brain-computer interface & 0.146 
& Detailed illustration of a futuristic brain-computer interface & 0.239 \\
& Blurred abstraction & 0.138 
& Detailed illustration of a futuristic biotechnology & 0.212 \\
& Detailed illustration of an alien world & 0.126 
& Detailed illustration of an alien world & 0.205 \\
& Illustration of an alien landscape & 0.126 
& Detailed illustration of a futuristic medical breakthrough & 0.202 \\
\midrule
\multirow{5}{*}{\rotatebox{90}{\textbf{Residuals}}} &
Detailed illustration of a futuristic brain-computer interface & 0.197
& Detailed illustration of a futuristic energy generator & 0.281 \\
& Fast-paced race car blur & 0.178
& Detailed illustration of a futuristic brain-computer interface & 0.278 \\
& Detailed illustration of a futuristic bioreactor & 0.164
& Detailed illustration of a futuristic virtual reality & 0.255 \\
& Illustration of an alien landscape & 0.162
& Detailed illustration of a futuristic medical breakthrough & 0.247 \\
& Detailed illustration of an alien world & 0.161
& Detailed illustration of a futuristic biotechnology & 0.244 \\
\midrule
  & \multicolumn{2}{c|}{\textbf{Stable Diffusion 1.4-200}} & \multicolumn{2}{c}{\textbf{DALL-E-1059}} \\\midrule
\multirow{5}{*}{\rotatebox{90}{\textbf{CLIP emb.}}} &
Vibrant city alley & 0.128
& Serene beach sunset & 0.126 \\
& Detailed illustration of an alien world & 0.116
& Surreal photo manipulation & 0.123 \\
& Photo featuring a vibrant street graffiti & 0.109
& Photo with grainy, old film effect & 0.118 \\
& Detailed illustration of a futuristic brain-computer interface & 0.092
& Photo with vintage film grain effect & 0.115 \\
& Photo featuring a vibrant urban graffiti & 0.090
& tranquil beach sunset & 0.111 \\
\midrule
\multirow{5}{*}{\rotatebox{90}{\textbf{Residuals}}} &
Vibrant city alley & 0.135
& Surreal photo manipulation & 0.147 \\
& Detailed illustration of a futuristic brain-computer interface & 0.134
& Detailed illustration of a futuristic brain-computer interface & 0.142 \\
& Photo featuring a vibrant street graffiti & 0.132
& Image with a futuristic nanobot swarm & 0.142 \\
& Detailed illustration of an alien world & 0.129
& Illustration of an alien landscape & 0.135 \\
& Photo with vibrant, saturated colors & 0.117
& Photo with grainy, old film effect & 0.131 \\
\bottomrule
\end{tabular}
\caption{...}
\label{tab:phrases_residuals}
\end{table*}

\begin{table*}[!t]\centering\setlength{\tabcolsep}{1.5pt}\footnotesize
\begin{tabular}{c|rl|rl|rl|rl|rl|rl } 
 \toprule
 \multicolumn{13}{c}{\textbf{Fake images produced by diffusion-based models}} \\\midrule
  & \multicolumn{2}{c}{$ $} & \multicolumn{2}{c|}{SD-1.4-200} & \multicolumn{2}{c}{DALL-E-mini} & \multicolumn{2}{|c}{GLIDE-base} & \multicolumn{2}{|c}{GLIDE-CLIP} & \multicolumn{2}{|c}{Kandinsky-100} \\\midrule
\multirow{5}{*}{\rotatebox{90}{CLIP embs.}} &
\multicolumn{3}{r}{gouache} & 0.08 & {\scriptsize instaweather} & 0.15 & blurred & 0.13 & blurred & 0.12 & uv & 0.12 \\ 
& \multicolumn{3}{r}{방탄소년} & 0.08 & 태 & 0.11 & blurry & 0.12 & blurry & 0.11 & fluor & 0.11 \\
& \multicolumn{3}{r}{bild} & 0.08 & webcamtoy & 0.10 & blur & 0.08 & 방탄소년 & 0.08 & renders & 0.10 \\
& \multicolumn{3}{r}{instaweatherpro} & 0.08 & gouache & 0.10 & 방탄소년 & 0.08 & octane & 0.08 & build & 0.10 \\
& \multicolumn{3}{r}{vscocam} & 0.07 & piccollage & 0.10 & octane & 0.08 & weil & 0.08 & busan & 0.09 \\\midrule
\multirow{5}{*}{\rotatebox{90}{Residuals}} &
   \multicolumn{3}{r}{bild} & 0.15 & scanned & 0.15 & meto & 0.12 & aku & 0.12 & bild & 0.16 \\ 
& \multicolumn{3}{r}{vscocam} & 0.11 & {\scriptsize instaweather} & 0.14 & bild & 0.11 & meto & 0.12 & uv & 0.16\\
& \multicolumn{3}{r}{방탄소년} & 0.09 & gouache & 0.14 & bungal & 0.11 & bild & 0.12 & renders & 0.15 \\
& \multicolumn{3}{r}{vsco} & 0.08 & webcamtoy & 0.14 & aku & 0.11 & bungal & 0.12 & fluor & 0.15 \\
& \multicolumn{3}{r}{pano} & 0.08 & 태 & 0.14 & 방탄소년 & 0.11 & weil & 0.11 & graphicdesign & 0.13 \\\midrule
\multicolumn{13}{c}{\textbf{Fake images produced by GAN-based models}} \\\midrule
& \multicolumn{2}{c}{GauGAN} & \multicolumn{2}{|c}{CycleGAN} & \multicolumn{2}{|c}{StyleGAN} & \multicolumn{2}{|c}{ProGAN} & \multicolumn{2}{|c}{BigGAN} & \multicolumn{2}{|c}{StarGAN} \\\midrule
\multirow{5}{*}{\rotatebox{90}{CLIP embs.}} & manatee & 0.15 & manatee & 0.14 & piccollage & 0.12 & piccollage & 0.14 & instaweather & 0.19 & {\scriptsize schwarzenegger} & 0.14 \\ 
& digital & 0.13 & {\scriptsize deeplearning} & 0.13 & stamatic & 0.09 & manatee & 0.12 & stamatic & 0.17 & transpa & 0.14 \\
& bungal & 0.12 & generative & 0.11 & {\scriptsize instaweather} & 0.09 & {\scriptsize instaweather} & 0.11 & png & 0.17 & oprah & 0.13 \\
& simulated & 0.12 & scanned & 0.10 & bharti & 0.06  & generative & 0.10 & piccollage & 0.16 & hani & 0.13 \\
& scrapped & 0.12 & effects & 0.10 & {\scriptsize instaweatherpro} & 0.06  & orthodon & 0.10 & {\scriptsize instaweatherpro} & 0.15 & zlatan & 0.13 \\\midrule
\multirow{5}{*}{\rotatebox{90}{Residuals}} & manatee & 0.16 & {\scriptsize deeplearning} & 0.16 & piccollage & 0.12 & manatee & 0.16 & generative & 0.17 & {\scriptsize schwarzenegger} & 0.15 \\ 
& generative & 0.14 & manatee & 0.15 & stamatic & 0.10  & generative & 0.14 & deeplearning & 0.16 & youngjae & 0.14 \\
& と繋 & 0.14 & generative & 0.13 & {\scriptsize instaweather} & 0.10  & neural & 0.13 & png & 0.16 & zlatan & 0.14 \\
& digital & 0.13 & generate & 0.13 & scanned & 0.08  & {\scriptsize deeplearning} & 0.13 & manatee & 0.16 & oprah & 0.14 \\
& octane & 0.13 & neural & 0.12 & simulated & 0.08  & neuro & 0.13 & digital & 0.15 & stephenking & 0.14\\
\bottomrule
\end{tabular}
\caption{Interpretation of LR weights $\w$ trained on fake/real data; each column shows 5 words whose CLIP embeddings are nearest to $\w$ and the corresponding cosine similarities.}\label{tbl:lrtokens_residuals}
\end{table*}

\subsection{Removing the components of CLIP embeddings}
\label{sec:hyperparams_removing}

Experiments with embeddings are performed in the following way. We consider images produced by generator number $i$ from our list and their real counterparts as one dataset, where generated and real images are labeled as ``0'' and ``1'' respectively. Then we use each image as input for the CLIP model and obtain the resulting embedding from the last layer of this model. Finally, we apply mean pooling to this embedding to reduce the dimensionality, obtaining a vector of dimension of $768$; this is our image feature vector.

We split the resulting dataset of these feature vectors into training, validation, and test subsets. Then we train logistic regression on the training subset and perform hyperparameter tuning on the validation subset, doing grid search upon the following sets of hyperparameters:
\begin{itemize}
    \item $\mathrm{solver} \in \{\mathrm{lbfgs}, \mathrm{saga}\}$;
    \item $L_2$ regularization coefficient \newline $C \in \{0.001, 0.005, 0.01, 0.05, 0.1, 0.5, 1, 10\}$;
    \item maximum number of iterations \newline $\mathrm{max\_iter} \in \{10, 50, 100, 500, 1000, 5000\}$.
\end{itemize}

After the best hyperparamters have been found on the validation subset, we merge train and validation subsets and train logistic regression on them. Then, we test the obtained classifier on the ``test'' subset of every other generator we have, and the resulting accuracy values are shown in the $i$th row of the diagram.
We repeat this process for every generator.


For experiments with residuals, we build classifiers based on only one feature constructed for every image, namely the dot product of the residual and (mean-pooled) embedding of the image.

\subsection{Accuracy plots}

\begin{figure}[!t]\centering
\includegraphics[width=\linewidth]{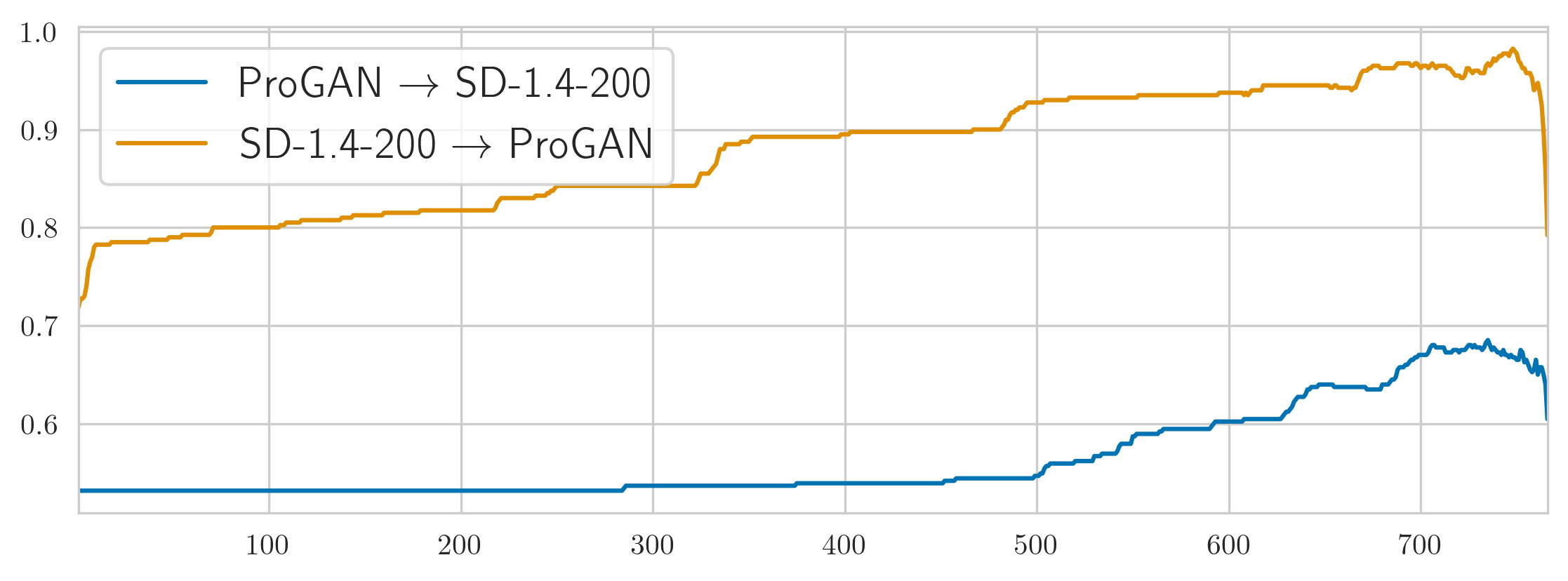}

\caption{Accuracy (vertical axis) as a function of the number of components removed from the CLIP-large embedding (horizontal axis), with the method described in Section~5.1. 
}\label{fig:progan_sd}
\end{figure}

The resulting scores for greedy feature search $S_{\mathrm{ProGAN} \to \mathrm{SD}}$ and $S_{\mathrm{SD} \to \mathrm{ProGAN}}$ are shown in Figure~\ref{fig:progan_sd}, as an example of typical score plots.




\subsection{Removing ``bad'' outliers and how it influences the geometry of embeddings}

\begin{figure}[!t]\centering
\includegraphics[width=.8\linewidth]{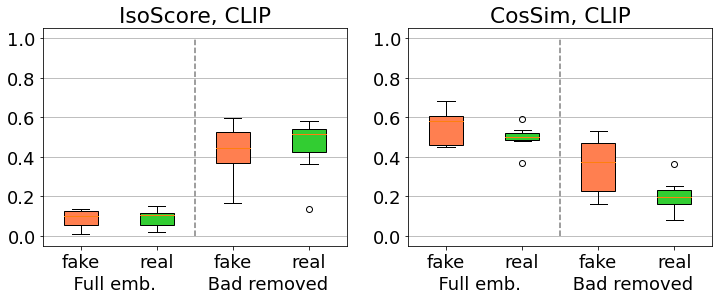}

\caption{IsoScore and cosine similarity of the CLIP-large embeddings before and after removing their ``bad'' components.}\label{fig:iso_CLIP}
\end{figure}

Some previous works have shown that some dimensions skew the embedding space greatly and have a dramatic influence on its geometry. In particular, \cite{timkey-van-schijndel-2021-bark} have shown that the embeddings of BERT, RoBERTa, and some other Transformer-based models lie in a narrow cone. To show this, they use the mean cosine similarity of the embeddings: if the cosine similarity of all embeddings is high, it means that they are similar to each other along some dimensions; the larger the average cosine similarity, the less isotropic the embedding space is.

\cite{rudman-etal-2022-isoscore} introduced a more complex tool for measuring the anisotropy of the embedding space: IsoScore. The fundamental motivation for IsoScore is that it roughly reflects the fraction of dimensions uniformly utilized by a given point cloud. According to the authors' estimation, less than 20\% of dimensions of the BERT model embedding space are utilized uniformly. Larger IsoScore values correspond to more isotropic embedding spaces.

Figure~\ref{fig:iso_CLIP} shows how removing the components that are ``bad'' for cross-domain and cross-model generalization abilities influences the IsoScore and cosine similarity scores for the CLIP-large feature extractor.
We see that after removing these ``bad'' dimensions, the embedding space of CLIP becomes more {isotropic}. Based on this result, we hypothesize that the isotropy of the embedding space can be connected to the model's generalization abilities; we leave testing this hypothesis for future research.

\clearpage

\end{document}